# EMC²A-Net: An Efficient Multibranch Cross-channel Attention Network for SAR Target Classification

Xiang Yu, Ying Qian, Zhe Geng, *Member*, IEEE, Xiaohua Huang, *Member*, IEEE, Qinglu Wang, and Daiyin Zhu *Fellow*, IEEE

*Abstract*—In recent years, convolutional neural networks (CNNs) have shown great potential in synthetic aperture radar (SAR) target recognition. SAR images have a strong sense of granularity and have different scales of texture features, such as speckle noise, target dominant scatterers and target contours, which are rarely considered in the traditional CNN model. This paper proposed two residual blocks, namely EMC²A blocks with multiscale receptive fields(RFs), based on a multibranch structure and then designed an efficient isotopic architecture deep CNN (DCNN), EMC²A-Net. EMC²A blocks utilize parallel dilated convolution with different dilation rates, which can effectively capture multiscale context features without significantly increasing the computational burden. To further improve the efficiency of multiscale feature fusion, this paper proposed a multiscale feature cross-channel attention module, namely the EMC²A module, adopting a local multiscale feature interaction strategy without dimensionality reduction. This strategy adaptively adjusts the weights of each channel through efficient one-dimensional (1D)-circular convolution and sigmoid function to guide attention at the global channel wise level. The comparative results on the MSTAR dataset show that EMC²A-Net outperforms the existing available models of the same type and has relatively lightweight network structure. The ablation experiment results show that the EMC²A module significantly improves the performance of the model by using only a few parameters and appropriate cross-channel interactions.

*Index Terms*—Channel attention, deep convolutional neural network (DCNN), isotopic architecture, multiscale feature fusion, synthetic aperture radar (SAR), target classification

## I. INTRODUCTION

SYNTHETIC aperture radar (**SAR**) imaging technology can obtain high-resolution images of geographical objects under all-day and all-weather conditions. Benefitting from the flexibility of its platform and the efficient remote sensing information acquisition capability, SAR imaging technology has wide application prospects in the military and civil fields.

Compared to optical images, a strong sense of granularity is an important characteristic of SAR images. In SAR target images, small texture features correspond to local details of the target and speckle noise, while the contour of the target often has larger texture features and usually has very high resolution. Due to these characteristics of the SAR image, the efficiency of manual interpretation of high-resolution SAR images is low. Therefore, SAR automatic target recognition (**SAR-ATR**) proposed by the MIT Lincoln Laboratory has received great attention.

Detection, discrimination and classification are three stages of SAR-ATR. Among them, SAR target classification has important strategic significance in the military field and attracts great interest. In classical SAR target classification methods, feature extraction [1-4] and classifier design [5-8] are two separate issues. Common used feature extraction methods include mathematical transformation features [1-2], computer vision features [3], electromagnetic features [4], etc. At present, the mainstream classifiers used in SAR-ATR include template matching [5], support vector machine [6], boosting [7], sparse representation [8], etc. However, traditional machine learning methods cannot automatically extract the features representing SAR targets. Therefore, in order to achieve this aim, some scholars have introduced the deep convolution neural network (DCNN) method, which can automatically extract the features of target images and enable end-to-end SAR target classification.

As a branch of deep learning theory, the application of DCNN in target classification and detection has received much attention and has achieved remarkable results. Nevertheless, there are still two major problems in applying the DCNN model for SAR target classification: (1) The intraclass difference of image features is large, and the number of samples is small. The SAR target image is very sensitive to the azimuth and pitch angles of images. Small changes in the above two angles may lead to drastic changes in the target backscattering coefficient and even the same targets in SAR images, which may have different features. Moreover, compared to the optical image dataset, the labelled samples of SAR target images are very scarce. Due to the large number of trainable parameters in the

This work was supported in part by Natural Science Foundation of Jiangsu Province under Grant BK20200420.

Xiang Yu, Ying Qian, Xiaohua Huang and Qinglu Wang, are with the School of Computer Engineering, Nanjing Institute of Technology, Nanjing 211167, China (e-mail: 419781300@qq.com).

Zhe Geng and Daiyin Zhu, are with the College of Electronic and information Engineering, Nanjing University of Aeronautics and Astronautics, Nanjing 211106, China.

*Corresponding author: Xiang Yu.*

network, the generalization ability of the DCNN model is greatly tested under a small-scale SAR target dataset. (2) Speckle noise interference. Due to small backscattering coefficient of natural scenery and the isotropic and uniform scattering, the background of SAR images has strong speckle noise [9]. The DCNN model is required to learn the local details of SAR targets in speckle noise, which intuitively both have small textural features, thus challenging the design of the DCNN model.

The concept of the local receptive field (**RF**) in CNNs comes from the study of neurons in the primary visual cortex of cats in the last century [10]. As convolution kernels with different RF sizes can collect multiscale spatial information in the same processing stage, this mechanism has been widely adopted in recent DCNNs. A typical application is the multibranch structure, which is composed of different RF convolution kernels in the same layer [11-13]. Relevant studies [9][14] have shown that convolution kernels with different RF sizes help improve the accuracy of SAR target recognition under speckle noise interference.

However, numerous experimental studies have suggested that the RF size of neurons in the visual cortex is not fixed but is modulated by the stimulus [15]. Inspired by this evidence, some nonlinear approaches (e.g., channel attention modules) [13-14] were proposed to aggregate information from adaptively selected features with different scales. In those channel attention modules, the weight activation core step was widely used to obtain the nonlinear channel weight, but some weaknesses persisted. In most current approaches, the channel weight activation is either within channels with the same RF [16-18] or between local channels with different RFs [13]. The mutual information between channels with the same and different RFs, which is called the multiscale RF cross-channel mutual information, does not receive much attention in the process of designing channel attention modules. It is believed that the channel weight could reflect the competitiveness of channel features in helping the model to achieve better performance. Therefore, the interaction of all feature maps in the same layer should be reflected not only in the feature fusion steps but also in the channel weight activation step.

Motivated by aforementioned studies, this paper designed a pair of novel and effective multibranch cross-channel attention (EMC$^2$A) blocks and further proposed a lightweight EMC$^2$A-Net. The main contributions of this work are as follows:

(1) Based on the multibranch structure, this paper proposed a pair of residual convolution blocks, namely, the EMC$^2$A block (A/B), with multiple RFs. EMC$^2$A blocks utilize parallel dilated convolution with different dilation rates to effectively capture multiscale texture features without significantly increasing the computational burden. The residual structure designed in the EMC$^2$A block (A/B) can alleviate the problem of gradient vanishing and exploding in training caused by block deep stacking.

(2) With the EMC$^2$A block (A/B), an efficient isotropic architecture DCNN EMC$^2$A-Net was designed. It is worth noting that the main part of EMC$^2$A-Net is just stacked by the EMC$^2$A block (A/B), so EMC$^2$A-Net can be designed flexibly according to the requirements of the task; such flexibility means that EMC$^2$A-Net has good expansibility.

(3) The EMC$^2$A module was proposed as a fully convolutional multiscale feature cross-channel attention module. In the EMC$^2$A module, to avoid side effects on channel attention prediction due to dimensionality reduction [19], a direct correspondence between the channel and its weight was developed. Moreover, benefitting from the multibranch cross-channel interaction structure in the EMC$^2$A module, the adaptively acquired channel weights reflected the global contribution of feature maps in both the scale and type dimensions. These characteristics make it efficient and lightweight.

The contrast and ablation experiments were carried out with the MSTAR SAR target dataset. **Without any data augmentation,** the contrast experiment results showed that EMC$^2$A-Net is superior to the previous state-of-the-art models (e.g., [13], [20]), and the total number of trainable parameters is relatively small. Then, based on EMC$^2$A-Net, the author demonstrated the effectiveness of the EMC$^2$A module through ablation experiments.

The rest of this paper is organized as follows. Section II gives an overview of the related work. Section III introduces the structure and components of the EMC$^2$A Net. Section IV discusses the dataset, implementations, and evaluation of the proposed method. Finally, conclusions are presented in Section V.

## II. RELATED WORKS

*A. Multibranch Convolutional Networks*

Convolutional network structures that contain more than one information transmission path are called multibranch convolutional networks. They are commonly categorized into the main path, combined with the bypass structure, and parallel paths with a multiscale RF structure.

The former aims at alleviating the vanishing and exploding gradient problems in training DCNNs. That was firstly introduced by highway networks [21] and then refined by ResNet [22], [23], while the bypassing path employs pure identity mapping. Furthermore, DenseNet [24] reflected this idea and added level wise multiple bypasses that enhance the integration of shallow features with high resolution and high-level features with rich semantic features.

The latter focuses on improving multiscale feature representations of layers in the network. For example, InceptionNets [25-28] carefully designed each branch with different RF convolution kernels, which not only improved the width of the network but also reduced the risk of overfitting. Such a structure with multi-RF kernels can indeed learn the multiscale texture features, but parallel two-dimensional (2D)-convolution operations with large kernels may consume too many computing resources during training and testing. Fortunately, dilated convolution can significantly reduce the computational cost and increase the RF of the convolution kernel simultaneously [29]. More recently, DINet was proposed for visual saliency prediction [12], which uses parallel dilated convolutions with different dilation rates in the same layer, making multiscale information extraction and fusion more efficient.

Noted that our proposed EMC$^2$A-Net follows the idea of Resnet and DINet; moreover, both identity mapping and multiscale feature fusion are integrated efficiently in EMC$^2$A-Net.

*B. Attention Mechanisms*

The attention mechanism has been indicated to be a potential enhancement means and has been applied in many research fields of DCNNs, such as semantic segmentation [30-32], object detection [33], [34] and target recognition [14], [35]. In the problem of target recognition, the attention mechanism can make some features that are conducive to improving the recognition rate more expressive and simultaneously suppress the features that contribute less.

SENet [16], as a milestone of a single-scale RF channel attention network, represents an efficient gating mechanism to self-recalibrate channel weights and achieved promising performance. BAM [17] and CBAM [18] entailed self-contained adaptive attention modules in both the spatial and channel axes.

With increased attention to the parallel multiscale RF structure, SKNet [13] followed the idea of SENet and introduced an adaptive channel attention mechanism into the multiscale RF kernel structure for the first time. However, further studies of ECA-Net show that channel wise dimensionality reduction in the above networks may have side effects on its weight prediction and then indicate that appropriate cross-channel interaction by using one-dimensional (1D) convolution is a more effective and efficient method for weight prediction [19]. Therefore, our designed EMC$^2$A module adopts 1D circular convolution to ensure the one-to-one mapping relationship between the channel and its weight. The multiscale RF cross-channel interaction is embodied in the design of the whole module, especially in the weight activation step, which was neglected in SKNet.

*C. Grouped/Shuffled/Dilated Convolution*

In the standard convolutional layer, the numbers of input and output channels (feature maps) are set to $C$ and $N$, respectively. When the kernel size is $K \times K$, and the total number of parameters of the kernel is $K \times K \times C \times N$; hence, the computational cost is expensive if $C$ is large. Group convolution reduced this cost, and the number of parameters was compressed to $K \times K \times (C/G) \times N$ by partitioning the input channels into $G$ mutually exclusive groups [36].

However, such group outputs from grouped convolution only relate to the inputs within the group, so the information interchange among groups is reduced. Fortunately, this cross-group information can be retained by the channel shuffle operation, which can also improve the performance of grouped convolution [37]. The reason is that the shuffle operation allows grouped convolution to obtain input data from different groups.

Dilated convolution was originally developed in a signal analysis algorithm for wavelet decomposition [38]. In the design of the DCNN, pooling and down sampling layers were commonly used to reduce image size and to obtain larger RFs. However, image structure information and spatial hierarchical information may be lost during these processes. To solve this problem, dilated convolution was introduced to enlarge the RF of the convolution kernel by increasing the dilation rate instead of reducing the image size. In DINet [12], "atrous spatial pyramid pooling (ASPP)", parallel dilated convolutions with different dilation rates were designed in the same layer, which can capture multiscale features and aggregate context information. In our designed EMC$^2$A-Net, dilated and grouped convolutions are used to improve the efficiency of capturing multiscale features, and channel shuffle operations with 1D circular convolution are also used to fuse the information among multiscale RF channels in the EMC$^2$A module.

III. OUR METHOD

For EMC$^2$A-Net, this paper adopted a 4-stage isotropic architecture and integrated a novel full convolution multibranch cross-channel attention module. In this architecture, the parallel multibranch structure was adopted, which can efficiently extract different scales of texture features from the same resolution feature map with fewer parameters. The lightweight network structure alleviated the overfitting problem in small-sample learning. In general, the contributions of feature maps to the recognition rate are different. To suppress the expression of features with less contribution (e.g., speckle features) and improve the expression of features with more contribution (e.g., target local features), our proposed EMC$^2$A module can adaptively adjust the weight of channels based on feature maps with different scales of features. In brief, the two contributions of this paper mainly focus on the design of the channel weight self-recalibration module, the EMC$^2$A module and the SAR target classification network, namely EMC$^2$A-Net. Next, the author introduces the EMC$^2$A module and network in two parts.

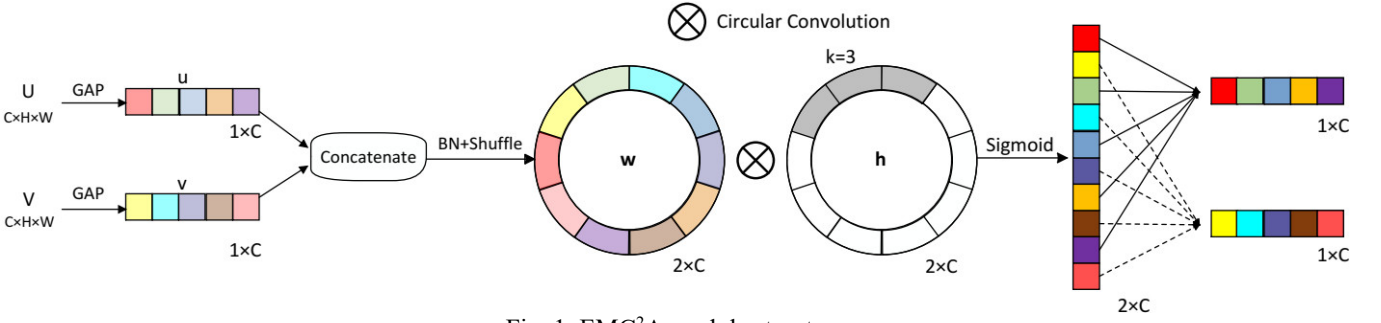

Fig. 1. EMC²A module structure

### A. EMC²A Module

*a. Compression*

The EMC²A module adopted multiscale features from a parallel multibranch convolution layer with different RFs. Its main structure is shown in Fig. 1. To simplify the description of the function of the module, only two groups of feature maps with two different feature scales, $U \in R^{C \times H \times W}$ and $V \in R^{C \times H \times W}$, are considered as two inputs to the module. Here, $C$, $H$, and $W$ are the number of channels and the vertical and horizontal sizes of the feature map, respectively. Since channel weights are determined by all feature maps themselves, it is necessary to reduce the dimensions of feature maps and keep the information independent of each other. For the above reasons, global average pooling (GAP) was used to compress the global information of each feature map into channel wise statistics. As shown in Fig. 1, the channel wise statistic vector $u$ ($1 \times C$) was calculated by shrinking U through spatial dimensions $H \times W$:

$$u = GAP(U) = \frac{1}{H \times W} \sum_{i=1}^{H} \sum_{j=1}^{W} U(i, j) \quad (1)$$

Similarly, the vector $v$ can be obtained.

*b. Shuffle and Fuse*

The main goal of the EMC²A module is to enable neurons to adaptively adjust their interests from multiple branches carrying different scales of features according to the stimulus content. To achieve this goal, we take two main steps: shuffle and fuse.

In the shuffle step, we need to gather all channel features and shuffle them. Specifically, we concatenate all GAP results, which are channel wise statistics of multiscale feature maps, and use a batch normalization (BN) operation to smooth the landscape of the whole loss function [39]. Then, all these channel wise statistics are shuffled so that channel information with multiscale features can be integrated into the next convolution layer.

For the problem of adaptive estimation of channel weights, the study in [20] indicates that avoiding channel wise statistical vector dimensionality reduction is helpful to learn effective channel attention, and it is inefficient and unnecessary to capture dependencies across all channels. We, inspired by this idea, proposed a channel wise statistics fusion and competition method in the fusion step for the parallel multiscale RF structure. In this method, a learnable and lightweight 1D-circular convolution was used. As shown in equation (2) and Fig. 1, $w$ is shuffled by channel wise statistics, $h$ is the learnable 1D-circular convolution kernel, and $R_N$ refers to the principal value interval with lengths $N$ and $N = 2 \times C$. Therefore, without padding, the dimension of the convolution result is $2 \times C$, which is consistent with the dimension of the input data.

$$y(n) = w(n) \otimes h(n) = \left[ \sum_{m=0}^{N} w(m) h((n-m))_N \right] R_N(n) \quad (2)$$

Let $k$ be the size of the convolution kernel, we used $R\_ncks$ to express the ratio of the number of channels to the kernel size of the 1D-circular convolution, which is verified and discussed in section IV part *E*. Cooperating with the previous shuffle step, the convolution kernel can perceive the channel wise statistic and its $k-1$ neighbors with multiscale features during each multiplication and addition operation. Therefore, the convolution result is an adaptive weighted average of channel wise statistics with multiscale features and reflects the interaction of all channels.

*c. Activation and Regrouping*

In the EMC²A module, to improve the performance of the model, all channels should adaptively adjust their weights through mutual competition during the network training process. Unfortunately, this is a nonmutually exclusive competitive relationship since multiple channels should be allowed to be emphasized (rather than enforcing a one-hot activation). To meet these criteria, we employed a gating mechanism with a sigmoid activation function to obtain the weights of all channels. Note that this activation operation makes full use of the mutual information of channels with multiscale RFs. Therefore, the channel weight integrates both the type and scale information of the feature maps in the EMC²A module.

The regroup operation is designed to reassign the channel weight vector back to two groups, which is actually an inverse operation of the shuffle step and ensures direct correspondence between the channel and its weight [20].

### B. EMC²A-Net

EMC²A-Net adopts a four-stage isotopic architecture, which is mainly composed of two kinds of blocks, namely, EMC²A Block-A and Block-B. Next, the author introduces EMC²A-Net from these two blocks.

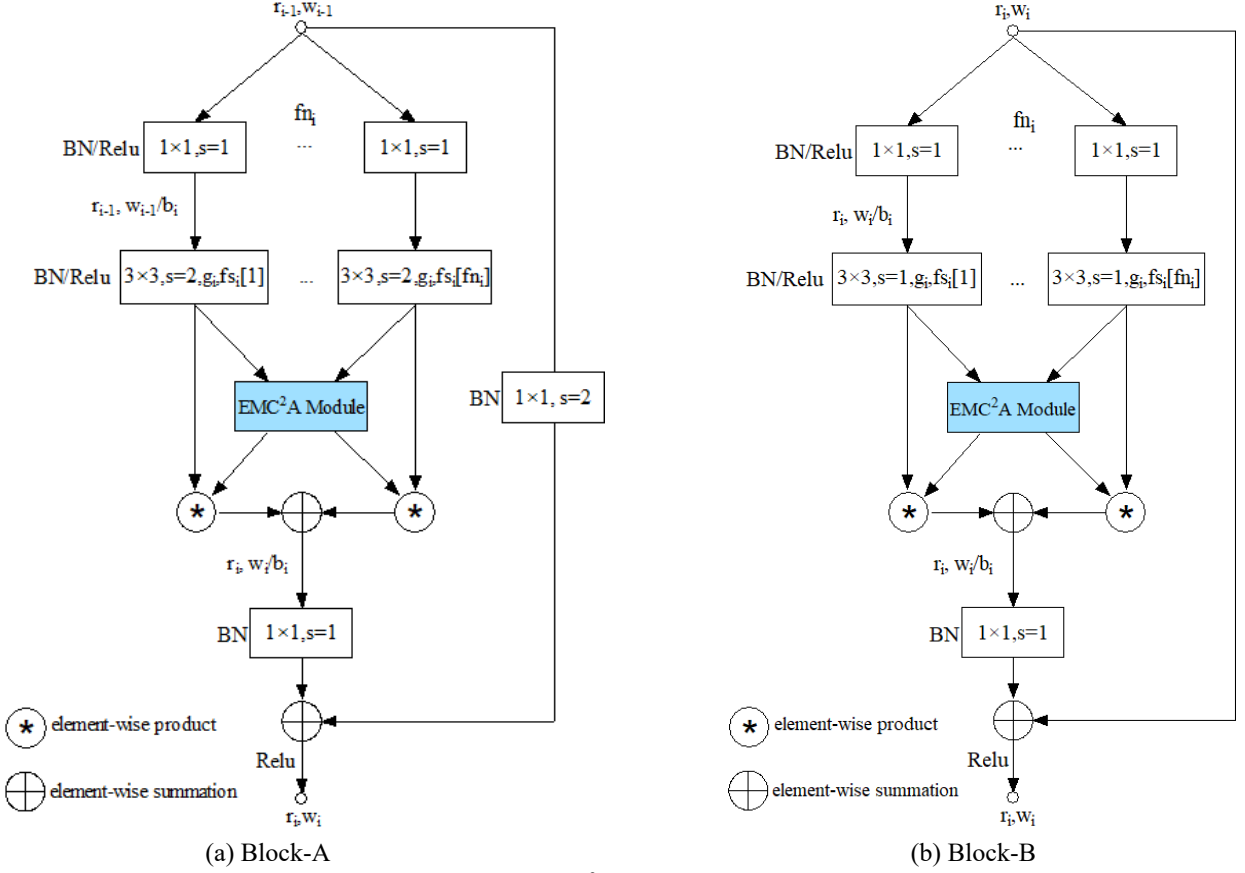

(a) Block-A  (b) Block-B

Fig. 2. EMC²A Block structure

*a. Blocks and Stages*

The structures of EMC²A Block-A and Block-B are shown in Fig. 2 where $r$ and $w$ are the size of the feature maps and number of channels, respectively, and the subscript $i$ represents the $i$ th stage. Both blocks adopt a structure like the classical ResBlock, but there are several differences. First, the $1 \times 1$ convolution layer does not change the number of channels $w_i$, which means that the bottleneck ratio is $b_i = 1$. This design increases the depth of the network and strengthens the network expression ability. Second, relevant studies [9], [14] have shown that multiscale features can improve the accuracy of SAR target recognition under speckle noise interference. Inspired by this idea, both blocks adopt multibranch and multiscale RF convolution structures based on the $3 \times 3$ dilated convolution, and the number of branches and the RF of each branch (dilation rate) are controlled by structural parameters $fn_i$ and $fs_i$, respectively. Another structure parameter $g_i$ is the number of groups in the group convolution. Third, the convolution output of all branches was fed into the EMC²A module to obtain channel weights, and the final fused feature maps were obtained by elementwise summation of weighted feature maps from each branch.

As seen from Fig. 2(a) and (b), Block-A and Block-B have two differences. First, in Block-A, the numbers of input and output channels ($w_{i-1}$ and $w_i$) are different. Due to a stride 2 ($s = 2$) and $3 \times 3$ convolution used in the Block_A main path, the sizes of the input and output feature maps (denoted as $r_{i-1}$ and $r_i$, respectively) are also different ($r_i = r_{i-1} / 2$). In contrast, for Block-B, its input and output have are the same dimension. Second, the skip connection can better transfer the gradient to the shallow layers in backpropagation, which can alleviate the problems of gradient vanishing and exploding to a certain extent. In Block-A and in Block-B, a stride ($s = 2$) $1 \times 1$ convolution and an identity mapping are used as the skip connection, respectively.

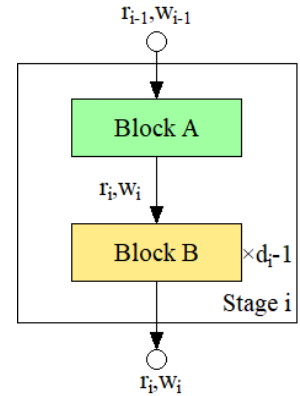

Fig. 3. Stage structure

Based on the above two blocks, we built the stage. As shown in Fig. 3. Block-A is used as the connection block between two stages. After the input feature map passed through Block-A, the size and channel number changed from $r_{i-1}$ and $w_{i-1}$ to $r_i$ and $w_i$, respectively. The main part of the stage is stacked by Block-B, and the number of Block-B is controlled by the structure parameter $d_i$.

*b. EMC²A-Net structure*

EMC²A-Net is a four-stage isotopic architecture DCNN that can be divided into three parts: stem, body and head, as shown in Fig. 4. In the stem part, a 3 × 3 convolution layer compresses the size of the input image and increase the number of channels. The body part in the middle is composed of four stages, as shown in Fig. 3. The structural parameters of each stage are specified in Table I, and when combined with the block structure (Fig. 2.), the body part of the network can be clearly described. The head part is composed of a fully connected layer and a softmax activation function, which are used as the final classification output. The cost function of EMC²A-Net adopts a cross-entropy function, and the weights of the network are updated by backpropagation according to the result of the cost function during the training process.

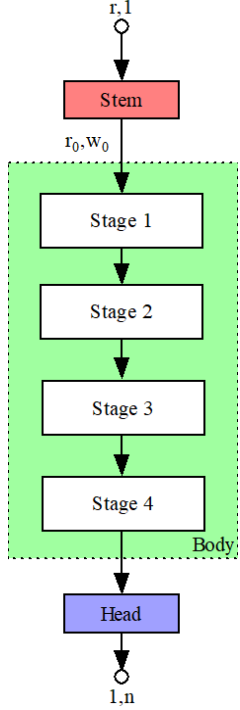

Fig. 4. EMC²A-Net structure diagram

TABLE I
STRUCTURAL PARAMETERS OF EACH STAGE

| Structural parameters | Stage 1 | Stage 2 | Stage 3 | Stage 4 |
|---|---|---|---|---|
| **d** | 2 | 3 | 2 | 1 |
| **w** | 128 | 64 | 200 | 256 |
| **b** | 1 | 1 | 1 | 1 |
| **g** | 16 | 8 | 25 | 32 |
| **fn** | 3 | 3 | 2 | 1 |
| **fs** | 1, 2, 3 | 1, 2, 3 | 1, 2 | 1 |

## IV. EXPERIMENTS

The experimental dataset used in this article is from the MSTAR program [40], which was launched in the mid-1990s and supported by the U.S. Defense Advanced Research Projects Agency (DARPA). The dataset is obtained by the STARLOS sensor, which is an X-band horizontal polarization SAR. The MSTAR program collected SAR images of various military vehicles of the former Soviet Union. Data included target occlusion, camouflage, configuration changes and other expansibility conditions, forming a more systematic and comprehensive measured database. Only a small subset of these data are publicly available on the website [41] and only the experimental dataset of this article are available. For simplicity, this subset is referred to as the MSTAR dataset below.

*A. Dataset and Implementation Details*

The MSTAR dataset includes ten different categories of military vehicles (armoured personnel carrier: BMP-2, BRDM-2, BTR-60, and BTR-70; tank: T-62, T-72; rocket launcher: 2S1; air defense unit: ZSU-234; truck: ZIL-131; bulldozer: D7) in spotlight mode within full aspect, and the resolution is 0.3 m*0.3 m. Examples of optical and SAR images of these targets at similar aspect angles are shown in Fig. 5.

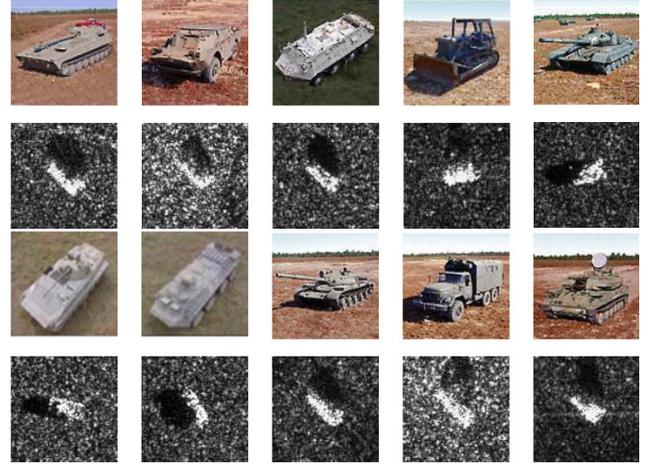

Fig. 5. Types of targets in the MSTAR dataset: (top) optical images versus (bottom) SAR images.

To comprehensively evaluate the performance of SAR target recognition algorithms, according to different acquisition conditions and target types, the MSTAR dataset is divided into three subsets: the standard operating condition (SOC) and extended operating conditions 1 and 2 (EOC-1 and EOC-2). In the SOC subset, the target types in the test and training sets are the same but with different aspects and depression angles. In the EOC subset, the difference between the training and test sets becomes even larger, including a more significant change in depression angle (EOC-1) or difference in the target series version (EOC-2). It is worth noting that a covariate shift problem was artificially introduced into the above three subsets. The covariate here refers to the input variables (images) of the model. Covariate shift refers to the input variables in the training and test sets having different data distributions [42]. A covariate shift in the dataset tests the extrapolation (a generalization ability) of the model. All experiments were performed on 2 pieces of Intel Xeon Gold 6240 CPUs and 1 Nvidia Tesla V100S GPU (32.0-GB). The images in the MSTAR dataset used in this article are in JPG format and were resized images to 158 × 158, except for A-ConvNet, whose input image size is 88 × 88, according to the original article. In the training process, the SGD optimizer was adopted, the learning rate (LR) adopted the cosine annealing descent method during 100 epochs, the initial LR was 0.005, and the final LR was 0.00001. In both the training and test processes, the batch size was set to 64. Data augmentation is not applied.

*B. Classification Results with SOC and Analysis*

The training and test sets of SOC contain 10 types of military vehicles that are completely consistent, while the imaging depression and azimuth angles are different, as shown in Fig. 5. The azimuth angle was evenly distributed between 0 and 360 degrees, while the depression angles of the training and test sets were 17° and 15°, respectively, as shown in Table II. It is well known that SAR image features are sensitive to the incident angle of radar beams. Therefore, there is a slight covariate shift between the training and test sets; this shift could be used to test the generalization ability of the models.

TABLE II
NUMBER OF TRAINING AND TEST IMAGES FOR THE SOC EXPERIMENTAL SETUP

| Types | Training set | | Test set | |
|---|---|---|---|---|
| | Depression | Number | Depression | Number |
| 2S1 | 17° | 299 | 15° | 274 |
| D7 | 17° | 299 | 15° | 274 |
| T62 | 17° | 299 | 15° | 273 |
| T72 | 17° | 232 | 15° | 196 |
| BRDM-2 | 17° | 298 | 15° | 274 |
| BTR-60 | 17° | 256 | 15° | 195 |
| BTR-70 | 17° | 233 | 15° | 196 |
| BMP-2 | 17° | 233 | 15° | 195 |
| ZIL-131 | 17° | 299 | 15° | 274 |
| ZSU-234 | 17° | 299 | 15° | 274 |

After training for 100 epochs, our proposed EMC$^2$A-Net achieved the highest accuracy of 99.7%. Fig. 6 demonstrates the confusion matrix of our proposed network, each row in the confusion matrix corresponds to the true category labels of the target, while each column represents the predicted category labels of targets. The main diagonal elements of the confusion matrix are significantly larger than those in other positions. That is, EMC$^2$A-Net achieved high classification accuracy in the SOC experiment.

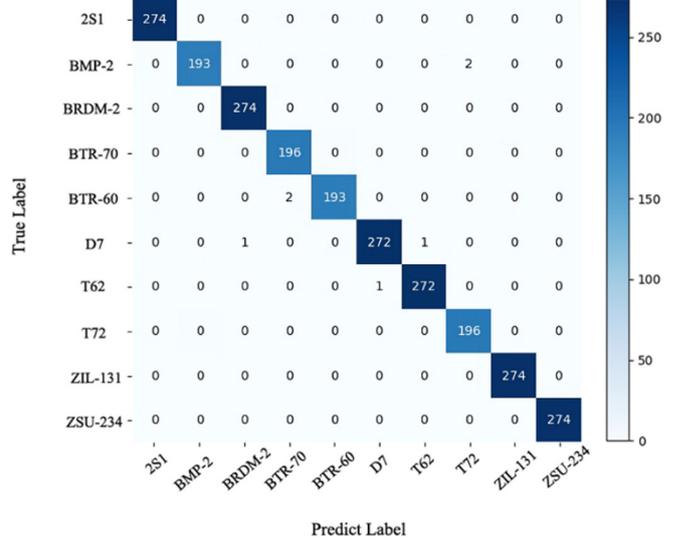

Fig. 6. Confusion matrix of the classification experimental results obtained by EMC$^2$A-Net with the SOC subset

To comprehensively analyze the performance of EMC$^2$A-Net, we also evaluated the 7 representative CNN models for comparison. The performance comparisons and computation time are reported in Table III, where the time consumed during the training (100 epochs) and testing (1 epoch) processes of eight models are calculated. The accuracy of these models exceeded 92%; among them, the accuracy of the DenseNet-121 model was relatively high, illustrating that learned image features can be strengthened by dense connections at different levels. EMC$^2$A-Net outperforms all models and has relatively smaller number of trainable parameters. Notably, EMC$^2$A-Net was comprehensively superior to SKNet and ResNext-50, which are also multibranch networks.

TABLE III
TEST RESULTS OF THE SOC SUBSET

| Networks | #Parameters (M) | Training time (s) | Test time (s) | Accuracy (%) |
|---|---|---|---|---|
| SENet | 1.23 | 289.55 | 0.15 | 92.8 |
| SKNet | 10.44 | 4338.52 | 0.47 | 93.4 |
| VGG-11 | 128.78 | 610.84 | 0.36 | 95.5 |
| A-ConvNet | 0.3 | 165.88 | 0.05 | 98.7 |
| ResNext-50 | 23 | 922.04 | 0.44 | 98.9 |
| ResNet-18 | 11.18 | 339.7 | 0.15 | 99.3 |
| DenseNet-121 | 6.96 | 803.35 | 0.65 | 99.4 |
| EMC$^2$A-Net | 0.96 | 901.46 | 0.42 | **99.7** |

*C. Classification Results with EOC and Analysis*

In the EOC-1 subset, the target categories of the test set and the training set are the same and contain four types of military vehicles. The distribution of the azimuth angle is like that of the SOC, which is between 0 and 360 degrees, while the depression angle (17° and 30°) gap between the training and test sets is much larger than that in the SOC, as shown in Table IV. Therefore, a significant covariate shift characteristic of the EOC subset will inevitably affect the test results.

TABLE IV
NUMBER OF TRAINING AND TEST IMAGES FOR THE EOC-1 EXPERIMENTAL SETUP

| Types | Training set | | Test set | |
|---|---|---|---|---|
| | Depression | Number | Depression | Number |
| 2S1 | 17° | 299 | 30° | 288 |
| T72 | 17° | 299 | 30° | 288 |
| BRDM-2 | 17° | 298 | 30° | 287 |
| ZSU-234 | 17° | 299 | 30° | 288 |

Interestingly, with the EOC-1 subset, EMC$^2$A-Net has still the highest accuracy of all the above models, although there is a slight decrease in accuracy (99.5%). As seen from the confusion matrix (Fig.7), misclassified samples are mainly concentrated between 2S1 and ZSU234. Compared with the data shown in Tables III and V, the test accuracy of the above models decreased with EOC-1, except for VGG-11, SENet and SKNet. In particular, DenseNet-121, the representative of the dense connection model, showed the largest decline in test accuracy. The above results show that increasing connections at different levels of the model cannot increase the generalization ability of the model, and the channel attention mechanism has more potential in combating the covariate shift problem. Since the number of samples in the training and test sets of EOC-1 is approximately 0.4 times the SOC, the training and test times are shorter.

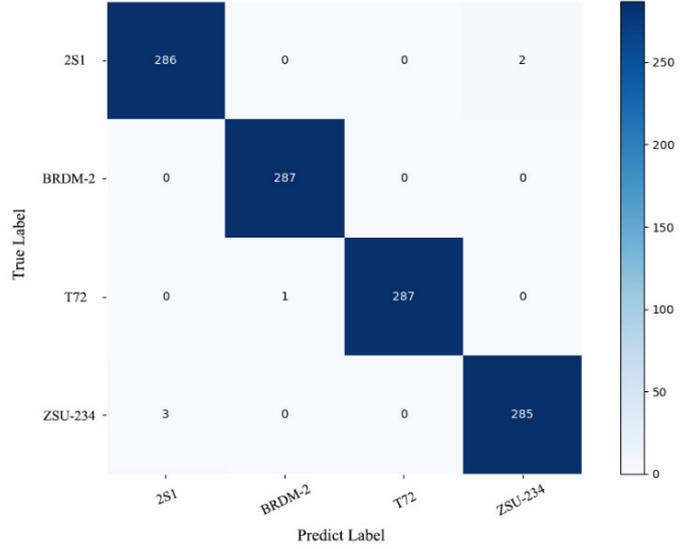

Fig. 7. Confusion matrix of the classification experimental results obtained by EMC$^2$A-Net with the EOC-1 subset

TABLE V
TEST RESULTS OF EOC-1

| Networks | #Parameters (M) | Training time (s) | Test time (s) | Accuracy (%) |
|---|---|---|---|---|
| DenseNet-121 | 6.96 | 421.67 | 0.25 | 91.8 |
| SENet | 1.23 | 215.13 | 0.07 | 95.8 |
| A-CONV Net | 0.30 | 132.15 | 0.04 | 96.2 |
| ResNext-50 | 23.00 | 454.82 | 0.16 | 98.7 |
| VGG-11 | 128.78 | 322.26 | 0.02 | 98.9 |
| SKNet | 10.44 | 1929.05 | 0.23 | 98.9 |
| ResNet-18 | 11.18 | 223.75 | 0.08 | 99.2 |
| EMC$^2$A-Net | 0.96 | 530.04 | 0.22 | **99.5** |

In the EOC-2 subset, as shown in Tables VI and VII, the depression angles of the training and test sets are partially different but relatively close. The training set contains four types of military vehicles, and the test set contains five series of T72 tanks that are not present in the training set. The appearance of different series leads to the distinction of their SAR image features in the details. The author regards the five series of T72 in the test set as the same classification (T72), and the preset four classifications of the test set are exactly the same as those of the training set. That is, there is only one ground-truth classification (T72) in the test set with four classifications. Overall, the covariate shift problem in the EOC-2 subset is more serious than that in the above two subsets.

TABLE VI
NUMBER OF TRAINING IMAGES FOR THE EOC-2 EXPERIMENTAL SETUP

| Types (Serial no.) | Training set | |
|---|---|---|
| | Depression | Number |
| BMP-2 | 17° | 233 |
| BRDM-2 | 17° | 298 |
| BTR-70 | 17° | 233 |
| T72(SN-132) | 17° | 232 |

TABLE VII
NUMBER OF TEST IMAGES FOR THE EOC-2 EXPERIMENTAL SETUP

| | Test set | | |
|---|---|---|---|
| Types | Serial no. | Depression | Number |
| T72 | SN-812 | 15°/17° | 426 |
| | SN-A04 | 15°/17° | 573 |
| | SN-A05 | 15°/17° | 573 |
| | SN-A07 | 15°/17° | 573 |
| | SN-A10 | 15°/17° | 567 |
| BMP-2 | / | / | 0 |
| BRDM-2 | / | / | 0 |
| BTR-70 | / | / | 0 |

The accuracy of our proposed EMC$^2$A-Net with EOC-2 is still acceptable at 95.3%. Fig. 8 illustrates that most of the predicted labels fall into the "T72" classification, and some misclassified samples are mainly distributed under "BMP-2" and "BRDM-2". As shown in Table VIII, the accuracy of all models tested with EOC-2 is significantly lower than previous accuracy measurements. It may be explained by that due to the serious covariate shift problem in the EOC-2 subset, some models cannot effectively distinguish similar characteristics of samples from some specific angles. In contrast, EMC$^2$A-Net and SKNet achieved relatively high accuracy (1st and 2nd) with EOC-2. More importantly, EMC$^2$A-Net is more efficient than SKNet. Interestingly, it is found that that both models are multibranch networks with a channel attention mechanism. Notably, the multibranch channel attention mechanism is a multiscale feature selection mechanism. In this mechanism, the strengthened features can alleviate the covariate shift problem of the dataset to a certain extent, and the weakened features can eliminate their negative impact on the covariate shift effect. The test with the EOC-2 subset shows that this mechanism plays a positive role in combating the covariate shift problem of the dataset.

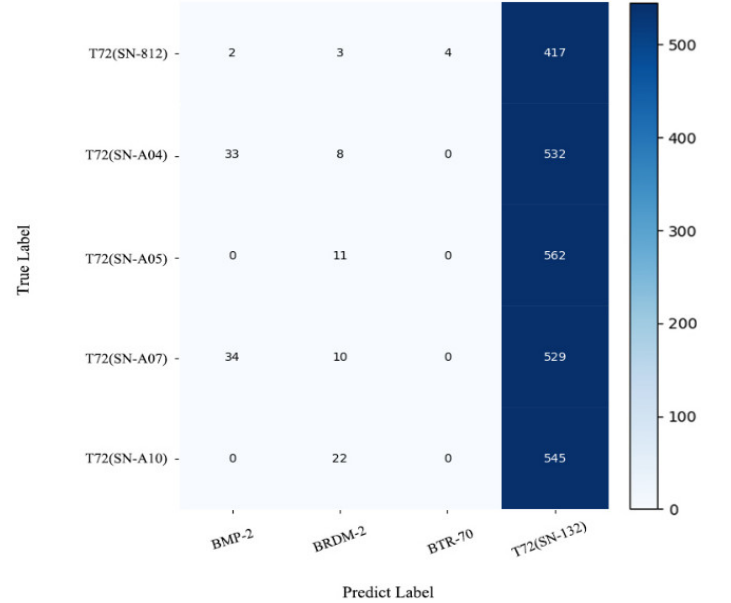

Fig. 8. Confusion matrix of the classification experimental results obtained by EMC$^2$A-Net with the EOC-2 subset

TABLE VIII
TEST RESULTS OF EOC-2

| Networks | #Parameters (M) | Training time (s) | Test time (s) | Accuracy (%) |
|---|---|---|---|---|
| DenseNet-121 | 6.96 | 453.60 | 0.73 | 78.5 |
| ResNext-50 | 23.00 | 538.95 | 0.41 | 82.2 |
| SENet | 1.23 | 205.90 | 0.18 | 83.1 |
| ResNet-18 | 11.18 | 225.30 | 0.14 | 87.1 |
| VGG-11 | 128.78 | 368.13 | 0.05 | 91.6 |
| A-CONV Net | 0.30 | 152.68 | 0.10 | 93.8 |
| SKNet | 10.44 | 2255.71 | 0.51 | 94.4 |
| EMC$^2$A-Net | 0.96 | 591.29 | 0.48 | **95.3** |

*D. Other Indicators*

In addition to test accuracy, precision, recall and F1-score [43] are also common evaluation indicators of dichotomous classification problems. They are defined as follows:

$$\text{Pricision} = \frac{\text{TP}}{\text{TP} + \text{FP}}, \quad (3)$$

$$\text{Recall} = \frac{\text{TP}}{\text{TP} + \text{FN}}, \quad (4)$$

$$\text{F1-Score} = \frac{2 \cdot \text{Precision} \cdot \text{Recall}}{\text{Precision} + \text{Recall}}. \quad (5)$$

In the above equation, TP, FP, and FN represent true positive, false positive, false negative, respectively. For the multiclass classification problem with the SOC and EOC-1 subsets, the macro-average is used to evaluate the performance of models, which are defined as follows:

$$\text{marco\_R} = \frac{1}{N} \cdot \sum_{j=1}^{N} \text{Recall}_j, \quad (6)$$

$$\text{marco\_P} = \frac{1}{N} \cdot \sum_{j=1}^{N} \text{Precision}_j, \quad (7)$$

$$\text{macro\_F1} = \frac{2 \cdot \text{macro\_P} \cdot \text{macro\_R}}{\text{macro\_P} + \text{macro\_R}}. \quad (8)$$

In Equations (6) and (7), N represents the total number of classifications. Subscript $j$ indicates classification $j$. The precision, recall and F1 score of each model are as described below.

TABLE IX
TEST RESULTS OF PRECISION, RECALL AND F1-SCORE FOR THE MSTAR DATASET

| Networks | SOC | | | EOC-1 | | | EOC-2 | |
|---|---|---|---|---|---|---|---|---|
| | macro_P | macro_R | macro_F1 | macro_P | macro_R | macro_F1 | Recall | F1-Score |
| ResNet-18 | 99.3 | 99.2 | 99.2 | 99.2 | 99.2 | 99.2 | 87.1 | 93.1 |
| ResNext-50 | 98.9 | 98.9 | 98.9 | 98.7 | 98.7 | 98.7 | 82.2 | 90.2 |
| DenseNet-121 | 99.3 | 99.3 | 99.3 | 93.5 | 91.8 | 91.6 | 78.5 | 87.9 |
| VGG-11 | 95.1 | 94.4 | 94.2 | 98.9 | 98.9 | 98.9 | 91.6 | 95.6 |
| SENet | 92.0 | 92.2 | 92.1 | 95.8 | 95.8 | 95.8 | 83.1 | 90.8 |
| SKNet | 96.3 | 96.2 | 96.2 | 98.9 | 98.9 | 98.9 | 94.4 | 97.1 |
| A-CONV Net | 98.7 | 98.7 | 98.7 | 96.2 | 96.2 | 96.2 | 93.8 | 96.8 |
| EMC2A-Net | **99.7** | **99.7** | **99.7** | **99.5** | **99.5** | **99.5** | **95.3** | **97.6** |

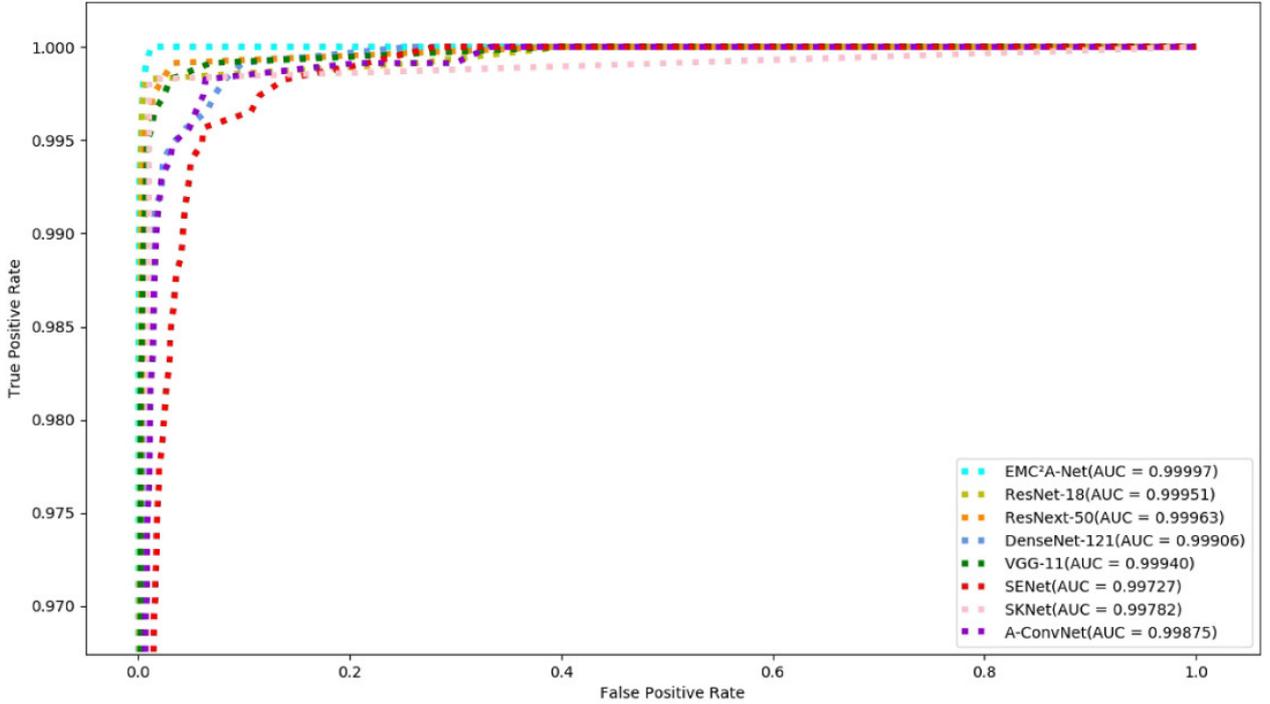

Fig. 9. ROC curves and the corresponding AUC values of different classification methods in the EOC-1 experiment.

Table IX shows that the precision, recall and F1-score indicators of EMC²A-Net are better than those of the other models. It is worth noting that the test set of EOC-2 contained four preset classifications, but in fact, there was only one classification, so only recall and F1 were calculated here.

The receiver operating characteristic (ROC) curves are an important metric for classification performance evaluation, which describe the relationship between the true positive rate (TPR) and the false positive ratio (FPR) by continuously altering the threshold value used in classification. TPR and FPR are defined as follows:

$$\text{TPR} = \frac{\text{TP}}{\text{TP} + \text{FN}} \quad (9)$$

$$\text{FPR} = \frac{\text{FP}}{\text{FP} + \text{TN}} \quad (10)$$

During the testing process of the experiment with the EOC-1 dataset, the ROC curves of the models mentioned above are shown in Fig. 9.

The results are shown in Fig. 9. The area under the ROC curve (AUC) indicators were adopted to assist in evaluating the performance of the above models. According to the definition of the ROC curve, models with larger AUCs usually have better performance. Compared with other models, the proposed EMC$^2$A-Net achieved a higher AUC and better performance.

*E. Ablation Experiment*

To objectively and comprehensively explain the advantages of the two main innovations in this paper, namely, EMC$^2$A-Net and the EMC$^2$A module, the author designed an ablation experiment for the SOC subset. The experiment was divided into two parts. In the first part, the author removed the EMC$^2$A module from EMC$^2$A-Net, subsequently called EMC$^2$-Net. Based on EMC$^2$-Net, we arranged experiments of single-scale RF and multiscale RF adaptation in four stages. Except for the dilation rate of the convolution kernel (fs) and the number of RF(fn), the other structural parameters of the network were the same. The results in Table X show that the parallel multiscale RFs based on the multibranch structure in EMC$^2$-Net are helpful for improving the recognition accuracy of SAR targets. However, it can also be seen that different scales of RF combinations make different contributions to improvements in accuracy.

TABLE X
THE TEST RESULTS OF THE PROPOSED MULTIBRANCH STRUCTURE WITH DIFFERENT COMBINATIONS OF FN AND FS IN EMC$^2$A-NET

| fn, fs | Stage 1 | Stage 2 | Stage 3 | Stage 4 | #Parameters (M) | Accuracy |
|---|---|---|---|---|---|---|
| fn$_i$, fs$_i$[1] | 1, 1 | 1, 1 | 1, 1 | 1, 1 | 0.638386 | 97.3 |
| fn$_i$, fs$_i$[1]~fs$_i$[fn$_i$] | 1, 1 | 2, 1, 2 | 1, 1 | 1, 1 | 0.669746 | 98.9 |
| fn$_i$, fs$_i$[1]~fs$_i$[fn$_i$] | 1, 1 | 2, 1, 3 | 1, 1 | 1, 1 | 0.669746 | 98.8 |
| fn$_i$, fs$_i$[1]~fs$_i$[fn$_i$] | 2, 1, 3 | 2, 1, 3 | 2, 1, 2 | 1, 1 | 0.873846 | 99.1 |
| fn$_i$, fs$_i$[1]~fs$_i$[fn$_i$] | 2, 1, 2 | 2, 1, 2 | 2, 1, 2 | 1, 1 | 0.873846 | 99.1 |
| fn$_i$, fs$_i$[1]~fs$_i$[fn$_i$] | 3, 1, 2, 3 | 3, 1, 2, 3 | 2, 1, 2 | 1, 1 | 0.964100 | **99.2** |
| fn$_i$, fs$_i$[1]~fs$_i$[fn$_i$] | 3, 1, 2, 4 | 3, 1, 2, 3 | 2, 1, 2 | 1, 1 | 0.964100 | 99.1 |

In the second part, based on the model with the best accuracy in Table IX, the author added the EMC$^2$A module, and the test results are shown in Table XI. Adding the EMC$^2$A module can help EMC$^2$-Net further improve the recognition accuracy of SAR targets, and the number of added parameters is quite small. In addition, as described in part *A(b)* of section III, the author tested the performance of different $R\_ncks$ with the EMC$^2$A module. Interestingly, the results show that with a certain number of channels, a smaller kernel size has a more significant effect on improving the performance of the EMC$^2$A module. Therefore, the author set the hyperparameter $R\_ncks$ to 32 and finally added the EMC$^2$A module to EMC$^2$-Net to form EMC$^2$A-Net.

TABLE XI
EMC$^2$A MODULE ABLATION ANALYSIS WITHIN EMC$^2$A-NET

| EMC$^2$A-module \ $R\_ncks$ | 4 | 8 | 16 | 32 | #Module parameters | Accuracy |
|---|---|---|---|---|---|---|
| - | | | | | 0 | 99.2 |
| √ | √ | | | | 782 | 99.5 |
| √ | | √ | | | 482 | 99.6 |
| √ | | | √ | | 330 | 99.6 |
| √ | | | | √ | **256** | **99.7** |

## V. Conclusions

In this paper, aiming to solve the problem of SAR target recognition, we proposed two residual blocks with multiscale RFs based on a multibranch structure and further designed a four-stage isotopic architecture DCNN, EMC$^2$A-Net, based on the above blocks. EMC$^2$A-Net has an efficient and lightweight structure. In addition, the author proposed a multiscale feature cross-channel attention module, the EMC$^2$A module, which adaptively generated channel weights by fusing the information of multiscale features from channels. This module improved the performance of the model with only a few parameters. Without any data augmentation technology, the results on the MSTAR dataset showed that compared with other classical and same-type target recognition models, EMC$^2$A-Net shows better performance and generalization ability.

REFERENCES

[1] Z. J. Liu, L. K. Zhuang, Y. F. Cao, *et al.*, "Target recognition of SAR images using principal component analysis and sparse representation," *Syst. Eng. Electron.,* vol. 35, no. 2, pp. 282–286, 2013.

[2] R. H. Huan and R. L. Yang, "Synthetic aperture radar images target recognition based on wavelet domain NMF feature extraction," *J. Electron. Inf. Technol.,* vol. 31, no. 3, pp. 588–591, 2009.

[3] Z. G. He, J. Lu, and G. Y. Kuang, "A survey on feature extraction and selection of SAR images," *Signal Process.,* vol. 24, no. 5, pp. 813–823, 2008.

[4] J. Lou, T. Jin, Q. Song, and Z.-m. Zhou, "Feature extraction of scattering centers in high-resolution SAR image," *J. Electron. Inf. Technol.,* vol. 33, no. 7, pp. 1661–1666, Aug. 2011, doi: 10.3724/sp.j.1146.2010.00960.

[5] T. D. Ross, S. W. Worrell, V. J. Velten, J. C. Mossing, and M. L. Bryant, "Standard SAR ATR evaluation experiments using the MSTAR public release data set," in *Proc. 3370 Algorithms Synthetic Aperture Radar Imagery V*, Orlando, FL: SPIE, 1998, pp. 566–573.

[6] Q. Zhao and J. C. Principe, "Support vector machines for SAR automatic target recognition," *IEEE Trans. Aerosp. Electron. Syst.,* vol. 37, no. 2, pp. 643–654, Apr. 2001, doi: 10.1109/7.937475.

[7] Y. Sun, Z. Liu, S. Todorovic, and J. Li, "Adaptive boosting for SAR automatic target recognition," *IEEE Trans. Aerosp. Electron. Syst.,* vol. 43, no. 1, pp. 112–125, Jan. 2007, doi: 10.1109/taes.2007.357120.

[8] G. Dong, G. Kuang, N. Wang, L. Zhao, and J. Lu, "SAR target recognition via joint sparse representation of monogenic signal," *IEEE J. Sel. Topics Appl. Earth Observ. Remote Sens.,* vol. 8, no. 7, pp. 3316–3328, Jul. 2015, doi: 10.1109/jstars.2015.2436694.

[9] Y. Gu and Y. Xu, "Architecture design of deep convolutional neural network for SAR target recognition," *J. Image Graph.,* vol. 23, no. 6, pp. 928–936, 2018.

[10] D. H. Hubel and T. N. Wiesel, "Receptive fields, binocular interaction and functional architecture in the cat's visual cortex," *J. Physiol.,* vol. 160, no. 1, pp. 106–154, Jan. 1962, doi: 10.1113/jphysiol.1962.sp006837.

[11] C. Szegedy *et al.*, "Going deeper with convolutions," in *2015 IEEE Conf. Comput. Vision Pattern Recognit. (CVPR)*, Boston, MA: IEEE, 2015, pp. 1–9.

[12] S. Yang, G. Lin, Q. Jiang, and W. Lin, "A Dilated inception network for visual saliency prediction," *IEEE Trans. Multimedia,* vol. 22, no. 8, pp. 2163–2176, Aug. 2020, doi: 10.1109/tmm.2019.2947352.

[13] X. Li, W. Wang, X. Hu, and J. Yang, "Selective kernel networks," in *2019 IEEE/CVF Conf. Comput. Vision Pattern Recognit. (CVPR)*, Long Beach, CA: IEEE, 2019, pp. 510–519.

[14] J. Ai, Y. Mao, Q. Luo, L. Jia, and M. Xing, "SAR target classification using the multikernel-size feature fusion-based convolutional neural network," *IEEE Trans. Geosci. Remote Sens.,* vol. 60, pp. 1–13, Sep. 2022, doi: 10.1109/tgrs.2021.3106915.

[15] J. I. Nelson and B. J. Frost, "Orientation-selective inhibition from beyond the classic visual receptive field," *Brain Res.,* vol. 139, no. 2, pp. 359–365, Jan. 1978, doi: 10.1016/0006-8993(78)90937-x.

[16] J. Hu, L. Shen, S. Albanie, G. Sun, and E. Wu, "Squeeze-and-excitation networks," *IEEE Trans. Pattern Anal. Mach. Intell.,* vol. 42, no. 8, pp. 2011–2023, Aug. 2020, doi: 10.1109/tpami.2019.2913372.

[17] J. Park, S. Woo, J.-Y. Lee, and I. S. Kweon, "Bam: Bottleneck Attention Module," *arXiv.org*, 18-Jul-2018. [Online]. Available: https://arxiv.org/abs/1807.06514. [Accessed: 29-Jul-2022].

[18] S. Woo, J. Park, J. Y. Lee, and I. S. Kweon, "CBAM: convolutional block attention module," in *Comput. Vision ECCV 2016*, V. Ferrari, M. Hebert, C. Sminchisescu, and Y. Weiss, Eds., Cham: Springer International Publishing, 2018, pp. 3–19.

[19] Q. Wang, B. Wu, P. Zhu, P. Li, W. Zuo, and Q. Hu, "ECA-Net: Efficient channel attention for deep convolutional neural networks," in *2020 IEEE/CVF Conf. Comput. Vision Pattern Recognit. (CVPR)*, Seattle, WA: IEEE, 2020, pp. 11531–11539.

[20] S. Chen, H. Wang, F. Xu, and Y. Q. Jin, "Target classification using the deep convolutional networks for SAR images," *IEEE Trans. Geosci. Remote Sens.,* vol. 54, no. 8, pp. 4806–4817, Aug. 2016, doi: 10.1109/tgrs.2016.2551720.

[21] R. K. Srivastava, K. Greff, and J. Schmidhube, "Highway networks," *arXiv:1505.00387,* May 2015, doi: 10.48550/arXiv.1505.00387.

[22] K. He, X. Zhang, S. Ren, and J. Sun, "Deep residual learning for image recognition," in *2016 IEEE Conf. Comput. Vision Pattern Recognit. (CVPR)*, Las Vegas, NV: IEEE, 2016, pp. 770–778.

[23] K. He, X. Zhang, S. Ren, and J. Sun, "Identity mappings in deep residual networks," in *Comput. Vision ECCV 2016*, B. Leibe, J. Matas, N. Sebe, and M. Welling, Eds., Cham: Springer International Publishing, 2016, pp. 630–645.

[24] G. Huang, Z. Liu, L. Van Der Maaten, and K. Q. Weinberger, "Densely connected convolutional networks," in *2017 IEEE Conf. Comput. Vision Pattern Recognit. (CVPR)*, Honolulu, HI: IEEE, 2017, pp. 2261–2269.

[25] C. Szegedy *et al.*, "Going deeper with convolutions," in *2015 IEEE Conf. Comput. Vision Pattern Recognit. (CVPR)*, Boston, MA: IEEE, 2015, pp. 1–9.

[26] S. Ioffe and C. Szegedy, "Batch normalization: Accelerating deep network training by reducing internal covariate shift," in *Proc. 32nd Int. Conf. Int. Conf. Mach. Learn.,* Lille, France: ICML, 2015, pp. 448–456.

[27] C. Szegedy, V. Vanhoucke, S. Ioffe, J. Shlens, and Z. Wojna, "Rethinking the inception architecture for computer vision," in *2016 IEEE Conf. Comput. Vision Pattern Recognit. (CVPR)*, Las Vegas, NV: IEEE, 2016, pp. 2818–2826.

[28] C. Szegedy, S. Ioffe, V. Vanhoucke, and A. Alemi, "Inception-v4, inception-ResNet and the impact of residual connections on learning," in *Proc. 31st AAAI Conf. Artif. Intell. (AAAI'17)*, California, CA: AAAI Press, 2017.

[29] P. Wang *et al.*, "Understanding convolution for semantic segmentation," in *2018 IEEE Winter Conf. Appl. Comput. Vision (WACV)*, Lake Tahoe, NV: IEEE, 2018, pp. 1451–1460.

[30] H. Liu, F. Liu, X. Fan, and D. Huang, "Polarized self-attention: Towards high-quality pixel-wise mapping," *Neurocomputing,* Jul. 2022, doi: 10.1016/j.neucom.2022.07.054.

[31] J. Fu et al., "Dual attention network for scene segmentation," in *2019 IEEE/CVF Conf. Comput. Vision Pattern Recognit. (CVPR)*, Long Beach, CA: IEEE, 2019, pp. 3141–3149.

[32] B. Zhang, J. Xiao, Y. Wei, K. Huang, S. Luo, and Y. Zhao, "End-to-end weakly supervised semantic segmentation with reliable region mining," *Pattern Recognition,* vol. 128, p. 108663, 2022.

[33] X. Dong, T. Gan, X. Song, J. Wu, Y. Cheng, and L. Nie, "Stacked hybrid-attention and group collaborative learning for unbiased scene graph generation," *arXiv:2203.09811,* Mar. 2022, doi: 10.48550/arXiv.2203.09811.

[34] S. Deng, X. Xu, C. Wu, K. Chen, and K. Jia, "3D AffordanceNet: A benchmark for visual object affordance understanding," in *2021 IEEE/CVF Conf. Comput. Vision Pattern Recognit. (CVPR)*, Nashville, TN: IEEE, 2021, pp. 1778–1787.

[35] X. Pan *et al.*, "On the integration of self-attention and convolution," *arXiv:2111.14556,* Nov. 2021, doi: 10.48550/arXiv.2111.14556.

[36] G. Huang, S. Liu, L. V. D. Maaten, and K. Q. Weinberger, "CondenseNet: An efficient DenseNet using learned group convolutions," in *2018 IEEE/CVF Conf. Comput. Vision Pattern Recognit.*, Salt Lake City, UT: IEEE, 2018, pp. 2752–2761.

[37] X. Zhang, X. Zhou, M. Lin, and J. Sun, "ShuffleNet: An extremely efficient convolutional neural network for mobile devices," in *2018 IEEE/CVF Conf. Comput. Vision Pattern Recognit.*, Salt Lake City, UT: IEEE, 2018, pp. 6848–6856.

[38] M. Holschneider, R. Kronland-Martinet, J. Morlet, and P. Tchamitchian, "A real-time algorithm for signal analysis with the help of the wavelet transform," in *Wavelets. Inverse Problems and Theoretical Imaging*, J. M. Combes, A. Grossmann, and P. Tchamitchian, Eds., Berlin, Heidelberg: Springer, 1990, pp. 286–297.

[39] S. Santurkar, D. Tsipras, A. Ilyas, and A. Madry, "How does batch normalization help optimization?" in *32nd Conf. Neural Inf. Process. Syst. (NIPS 2018)*, Montréal, CA, 2018, pp. 1–11.

[40] E. R. Keydel, S. W. Lee, and J. T. Moore, "MSTAR extended operating conditions: A tutorial," in *Proc. 3rd SPIE Conf. Algorithms SAR Imagery*, Orlando, FL: SPIE, 1996, pp. 228–242.

[41] "The Air Force Moving and Stationary Target Recognition Database." [Online]. Available: https://www.sdms.afrl.af.mil/datasets/mstar/.

[42] D. Y. Geeta, N. G. Nair, P. Satpathy, and J. Christopher, "Covariate shift: A review and analysis on classifiers," in *2019 Global Conf. Advancement Technol. (GCAT)*, Bangalore, India: IEEE, 2019, pp. 1–6.

[43] F. Yang, Q. Xu, and B. Li, "Ship detection from optical satellite images based on saliency segmentation and structure-LBP feature," *IEEE Geosci. Remote Sens. Lett.,* vol. 14, no. 5*,* pp. 602–606, May 2017, doi: 10.1109/lgrs.2017.2664118.

AUTHOR BIOGRAPHIES

**Xiang Yu** received the Ph.D. degree in communication and information system from the Nanjing University of Aeronautics and Astronautics, Nanjing, in 2014. He is currently an Associate Professor with Nanjing Institute of Technology, Nanjing, China. His research interests include SAR/ISAR imaging, radar target detection and embedded system design.

**Qian Ying** received the master degree in system engineering from Nanjing University of Science & Technology, Nanjing, China, in 2006.She is currently a Lecturer in the School of Computer Engineering, Nanjing Institute of Technology, Nanjing, China. Her main research interest is Image Processing, Computer Vision and Pattern Recognition.

**Zhe Geng** received the Ph.D. degree in electrical engineering from Florida International University, Miami, FL, USA, in 2018. From 2018 to 2019, she was a research scientist with Wright State University, Dayton, OH, USA. In Dec. 2019, Dr. Geng jointed the College of Electronic and Information Engineering, Nanjing University of Aeronautics and Astronautics, Nanjing, China, where she is currently a Research Associate. Her research interests include deep learning, image processing, and automatic target recognition.

**Xiaohua Huang** received the B.S. degree in communication engineering from Huaqiao University, Quanzhou, China in 2006. He received his Ph.D degree in Computer Science and Engineering from University of Oulu, Oulu, Fin- land in 2014. He was a research assistant in Southeast University since 2006. He had been a senior researcher in the Center for Machine Vision and Signal Analysis at University of Oulu in 2015-2019. He had also been research associate at University of Cambridge. He is currently a professor at Nanjing Institute of Technology, China, and distinguished professor of Jiansu province. He has authored or co-authored more than 40 papers in journals and conferences, and has served as a reviewer for respective journals and conferences. His current research interests include emotion recognition, object detection, handwritten recognition and texture classification. He is a member of the IEEE.

**Qinglu Wang** was born in Fujian, in 2001. He is currently pursuing a bachelor's degree at Nanjing Institute of Technology, Jiangsu, China.
His research interests include remote sensing image processing and artificial intelligence

**Daiyin Zhu** was born in Wuxi, China, in 1974. He received the B.S. degree in electronic engineering from the Southeast University, Nanjing, China, in 1996 and the M. S. and Ph. D. degrees in electronic from the Nanjing University of Aeronautics and Astronautics (NUAA), Nanjing, in 1998 and 2002, respectively.
From 1998 to 1999, he was a Guest Scientist with the Institute of Radio Frequency Technology, German Aerospace Center, Oberphaffenhofen, where he worked in the field of SAR interferometry. In 1998, he joined the Department of Electronic Engineering, NUAA, where he is currently a professor. He has developed algorithms for several operational airborne SAR systems. His current research interests include radar imaging algorithms, SAR/ISAR autofocus techniques, SAR ground moving target indication (SAR/GMTI), and SAR interferometry.